\documentclass[conference]{IEEEtran}
\IEEEoverridecommandlockouts
\usepackage{cite}
\usepackage[hidelinks,colorlinks=true,linkcolor=blue,citecolor=blue]{hyperref}
\usepackage{amsmath,amssymb,amsfonts}
\usepackage{algorithmic}
\usepackage{graphicx}
\usepackage{textcomp}
\usepackage{xcolor}
\usepackage{amsmath}
\usepackage{subcaption}
\usepackage{flushend}
\usepackage{graphicx}
\usepackage{lipsum}
\usepackage{booktabs}
\begin{document}

\title{End-to-End Multi-View Structure-from-Motion with Hypercorrelation Volumes
}

\author{\IEEEauthorblockN{Qiao Chen, Charalambos Poullis}
\IEEEauthorblockA{\textit{Immersive and Creative Technologies Lab} \\
\textit{Department of Computer Science and Software Engineering, Concordia University} \\
Montreal, Quebec, CA}
}

\maketitle

\begin{abstract}
Image-based 3D reconstruction is one of the most important tasks in Computer Vision with many solutions proposed over the last few decades. The objective is to extract metric information i.e. the geometry of scene objects directly from images. These can then be used in a wide range  of applications such as film, games, virtual reality, etc. Recently, deep learning techniques have been proposed to tackle this problem. They rely on training on vast amounts of data to learn to associate features between images through deep convolutional neural networks and have been shown to outperform traditional procedural techniques. In this paper, we improve on the state-of-the-art two-view structure-from-motion(SfM) approach of \cite{wang2021deep} by incorporating 4D correlation volume for more accurate feature matching and reconstruction. Furthermore, we extend it to the general multi-view case and evaluate it on the complex benchmark dataset DTU \cite{jensen2014large}. Quantitative evaluations and comparisons with state-of-the-art multi-view 3D reconstruction methods demonstrate its superiority in terms of the accuracy of reconstructions.

\end{abstract}


\section{Introduction}
3D perception is an important ability of the visual system that can improve scene understanding. With the rapid development of computer technology, 3D reconstruction techniques are playing an increasingly important role in all aspects of industry and production, such as preservation of cultural heritage, virtual reality and other fields. Perhaps the most popular and successful 3D reconstruction technique in recent years is based on structure-from-motion and multi-view stereo. This is due to the low equipment cost, high operational flexibility, and good reconstruction accuracy it provides.

Recently, deep learning models have been proposed to address vision tasks including 3D depth estimation from images. Given a large amount of annotated data, a model can learn the nonlinear mapping between source and target domains. It has been shown that deep learning models can also perform feature detection, pose estimation, landmark localization, and image recognition tasks. The results achieved are remarkable; in many problems related to expression learning, scholars have also applied it to multi-view learning, resulting in a number of works on multi-view stereo  such as \cite{yao2018mvsnet,yao2019recurrent}. 

In this paper, we provide the research background on multi-view reconstruction and explore related concepts, including an overview of 3D reconstruction methods, the concept of multi-view features, and structure-from-motion reconstruction and multi-view stereo. Next, we present an end-to-end model for multi-view structure-from-motion with hypercorrelation volume which results in high accuracy feature matching and improved 3D reconstructions. Lastly, we evaluate our technique on the complex multi-view stereo benchmark dataset DTU\cite{jensen2014large} and present a quantitative evaluation and comparison with state-of-the-art 3D reconstruction methods demonstrating the superiority of our approach.

\section{Related Work}
\subsection{3D reconstruction}
3D reconstruction has experienced a long process of improvement, produced rich results, and played an important role in life and production nowadays. In some traditional fields, such as in industrial manufacturing, reverse engineering based on 3D reconstruction technology can help producers effectively improve product quality; for example, in cultural heritage protection, building 3D models is the digital preservation of cultural relics. In some emerging fields, such as virtual reality, high-precision 3D reconstruction technology brings users a more immersive experience. In addition, 3D reconstruction is also widely used in the fields of smart medical care, urban planning, and autonomous driving. With the continuous improvement of science and technology, the application scope and application requirements of 3D reconstruction are also expanding. Therefore, it is particularly important to study accurate and robust 3D reconstruction algorithms. 

3D reconstruction can be divided into active and passive reconstruction based on the way of acquiring 3D models. Active reconstruction is scanning the object from all directions using a three-dimensional scanning device, and directly obtaining the three-dimensional model of the object to be reconstructed. Using the energy source emitted by the 3D scanning device itself, the 3D scanning device mainly refers to a laser rangefinder or a structured light scanner. Structured light scanners have lower requirements for the scanning environment, and the scanning results can be used as auxiliary tools for other applications. However, active reconstruction using structured light has its disadvantages. For example, it cannot directly and accurately obtain texture information related to the application scene, and it cannot be directly applied to a wide range of open scenes. Moreover, the cost while using the 3D scanning equipment, as well as the price of the equipment itself, is also very expensive. It is only used in specific situations, and the active reconstruction has more limitations. In passive reconstruction, images are obtained directly through the traditional RGB cameras, and then a 3D reconstruction is performed on the images using algorithms such as structure-from-motion. In the actual application process, passive reconstruction directly uses the camera to capture the image, performs feature matching between the images, and then triangulates to calculate the depth information of the reconstructed scene to restore the three-dimensional structure of the objects or the scene. Compared with active reconstruction, passive reconstruction relies on calculating 3D points rather than measuring them, it does not require expensive acquisition equipment and has the advantages of a large measurement range and high variability, which is convenient for manual operation. The simplest method only requires cameras and computers.

\subsection{Structure-from-Motion}

Structure-from-Motion(SfM) is a feature-based 3D scene reconstruction algorithm in the field of multi-view geometry reconstruction in computer vision. The algorithm recovers camera parameters and scene structure information by analyzing the motion process of the camera relative to the target scene. By inputting images of a target from different perspectives, the overall three-dimensional structure of the scene is recovered. At present, the algorithm is mainly used to recover the camera motion trajectory through video successively tilt or recover the three-dimensional structure information of the scene through the multi-view picture set of the same scene.
Snavely et al \cite{agarwal2011building} and Furukawa et al \cite{furukawa2010towards} proposed the modern structure-from-motion and dense patch matching reconstruction from unstructured 2D images. Schonberger et al \cite{schonberger2016structure} revisited this problem and proposed a state-of-the-art procedural structure-from-motion pipeline, which is widely used in other applications and even used to generate ground truth in deep learning approaches.

\subsection{Deep Learning Multi-View 3D Reconstruction}

With recent active research on deep learning, an increasing number of neural network-based 3D reconstruction methods have been proposed. Most deep learning neural networks employ a convolutional neural network(CNN) \cite{NIPS2012_4824} and can be categorized into two types. In the first category\cite{zhou2017unsupervised}, similar to structure-from-motion, the problem can be summarized as a joint optimization task of monocular depth and camera poses; in the second category\cite{yao2019recurrent}, similar to multi-view stereo, the camera poses are known and the depths are iteratively refined via multi-view geometry. The main difference is the recovered 3D information density and multi-view stereo techniques perform reconstruction on implicit surfaces and radiance fields.

In this paper, we extend a deep two-view structure-from-motion method to the more general task of multi-view reconstruction. Different from the two-view camera and depth estimation, the multi-view stereo tasks require both, camera recovery and depth estimation. In this work, we show deep learning-based multi-view reconstruction on a complex dataset i.e. DTU benchmark dataset\cite{jensen2014large} and carry out the comparison of the 3D models between this approach and state-of-the-art multi-view stereo techniques. 

\begin{figure*}[!ht]
\centering
\includegraphics[width=0.87\textwidth]{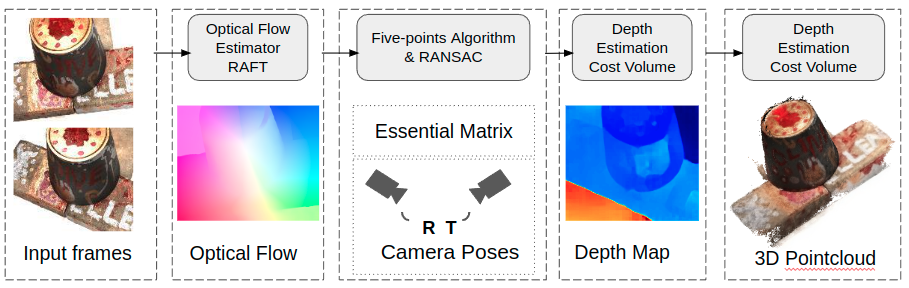}
\caption{Overview of the 3D reconstruction pipeline.}
\label{fig:system_overview}
\end{figure*}
\setlength{\textfloatsep}{10pt plus 1.0pt minus 5.0pt}

\section{Methodology}

The proposed framework follows the classic pipeline for 3D reconstruction from multi-view images where the features in images are matched, followed by the camera pose and the relative depth map estimation. In the last step, homography warping generates the final output. The pipeline is shown in Figure \ref{fig:system_overview}. Below, we explain the two main steps: (i) the structure-from-motion based on deep learning that results in the relative depth maps between pairs of images, and (ii) multi-view reconstruction with homography warping which results in the final pointcloud.  

\subsection{Structure-from-Motion with Deep Learning model}
Structure-from-motion involves the three main steps, feature matching, camera recovery, and depth map estimation.

\subsubsection{Feature Matching}
The first step is matching features between pairs of image frames. Unlike traditional techniques which rely on sparse feature correspondence, we employ an optical flow estimation network that predicts dense correspondences between pairs of images. Deep learning models for optical flow have been shown to generate per-pixel dense matches that are robust to different textures, occlusions, large displacements and deformations. We integrate the state-of-the-art deep optical flow network RAFT\cite{raft2021-662} to generate dense feature matches. The RAFT architecture extract per-pixel features, it employs multi-scale 4D hypercorrelation volumes, and iteratively updates the flow field through a recurrent unit. The model is pretrained on FlyingChairs\cite{dosovitskiy2015flownet} and FlyingThings\cite{mayer2016large}, followed by fine-tuning on the benchmark dataset, e.g. MPI-Sintel\cite{butler2012naturalistic} and achieves an end-point error (EPE) of 2.855 pixels, one of the lowest in recent years.

Given a pair of consecutive RGB images $I_1$ and $I_2$, features are extracted from the input images using a convolutional network. The feature encoder and context encoder extracts per-pixel features. The encoder network is applied to both images and maps them to dense feature maps. The correlation layer computes the visual similarity between pixels. Given the image features maps,  the hypercorrelation volume is formed by calculating the cosine similarity between all pairs of per-pixel feature vectors. A 4-layer pyramid is constructed by pooling the last two dimensions of the correlation volume with kernel sizes 1, 2, 4, and 8, followed by lookups on all levels of the pyramid to compute each correlation value. Lastly, the update operator imitates the process of an iterative optimization, with the iterative updates, the update operator estimates a sequence of flow estimation and produces an update direction that applies to the current estimation. The update operator takes the flow and correlation as an input and outputs the update direction until its convergence to a fixed point. The update operator works as an energy minimization and optimization function which outputs the final optical flow estimation.

\subsubsection{Camera Recovery}

The normalized pose estimation module computes relative camera poses from the 2D optical flow correspondences. Given a set of matching points and the camera intrinsic matrix $K$, the essential matrix $E$ can be recovered from the five-point algorithm\cite{nister2004efficient}. Given a pair of matches $\boldsymbol{x}_i$ and $\boldsymbol{x'}_i$ and the  camera intrinsic matrix $K$, structure-from-motion finds a camera rotation matrix $R$ and a translation vector $T$. The point $X_i$ is the result of the triangulation of the corresponding points given by,
\noindent
\begin{align*}\label{eq:cam}   
  & \boldsymbol{x}_i = K [I | 0]Xi 
  & \boldsymbol{x'}_i = K[R | T]Xi 
\end{align*}
where $I$ is the identity matrix. To recover the essential matrix $E$ we need at least 5 points, followed by $R$ and $T$ decomposition from $E$. After the triangulation of the matched points, $x_i$ and $x'_i$ using the camera poses, we get the 3D points represented as a depth map for each pair of images in the image sequence.

\subsubsection{Depth Estimation}
Although the camera recovery results in dense depth maps, the depth estimation process samples a subset of these otherwise it would not take advantage of the epipolar constraint and re-calculates the depth. This process is similar to multi-view stereo, where the standard plane-sweep algorithm samples the distribution of matching candidates and estimates the depth. We train the model in an end-to-end manner and supervise it using ground truth depth maps. Moreover, to make the depth estimation scale-invariant, the translation vectors are normalized and the matching candidates are depending on the camera poses and the scale factor. The loss function is given by,
\noindent
\begin{align*}
   \boldsymbol{L}_{depth} &= \sum_{\boldsymbol{x}} l_{huber}(\hat{\boldsymbol{d_x}} -\boldsymbol{d_x})
   \\
   \boldsymbol{L}_{flow} &= \sum_{\boldsymbol{x}} (\hat{\boldsymbol{o_x}} -\boldsymbol{o_x})^2\\
   \boldsymbol{L}_{total} &= \boldsymbol{L}_{depth} + \boldsymbol{L}_{flow}
\end{align*}
where $\hat{\boldsymbol{d_x}}$ is the predicted depth and
$\boldsymbol{d_x}$ is the ground truth depth, $\hat{\boldsymbol{o_x}}$ is the predicted optical flow and $\hat{\boldsymbol{o_x}}$ is the ground truth optical flow. The total loss $\boldsymbol{L}_{total}$ can be calculated from the sum of the depth loss $\boldsymbol{L}_{depth}$ and optical flow loss $\boldsymbol{L}_{flow}$, the Huber function is given by,
\[
l_{huber} (z)= 
\begin{cases}
    0.5z^2, \text{if } \mid z \mid < 1\\
    \mid z-0.5 \mid,  \text{otherwise}
\end{cases}
\]
where $z$ is the estimated depth.
\vspace{-5pt}
\subsection{Multi-view 3D Reconstruction}
\vspace{-5pt}
Reconstruction of the scene geometry is the last step. Given the pair-wise depth maps, we merge the reconstructions into a global point cloud using homography warping which implicitly uses camera geometries to build the 3D volume. To generate the 3D point cloud from the per-view depth maps, we apply depth map fusion, which integrates depth maps from different views to a unified point cloud representation. In the visibility-based fusion algorithm \cite{yao2018mvsnet}, depth occlusions and violations across different viewpoints are minimized and the average overall reprojected depth is taken as the final depth estimation of the pixel. Finally, the fused depth maps are directly reprojected to generate the 3D point cloud.

\section{Experiments}
\vspace{-5pt}

We implemented our framework in Python3.8 with Pytorch 1.6. All experiments were conducted on a workstation with an Intel i7 processor and a NVIDIA GTX 3080Ti graphics card.

\subsection{Dataset}

The dataset used in this paper is the benchmark dataset DTU presented in \cite{jensen2014large}. The DTU dataset is a large-scale multi-view 3D reconstruction dataset collected in a strictly controlled laboratory environment. The real-world annotations provided by this dataset are the point cloud data collected by the structured light scanner, which allows for the quantitative evaluation of the results of the 3D reconstruction. The dataset consists of 124 different scenes rotated and scanned four times at 90-degree intervals, hence giving a complete view of the models. Each scene was captured with 49 or 64 images with a resolution of $1600 \times 1200$. 

\subsection{Quantitative Analysis }
In this section we present a quantitative evaluation and comparison between the proposed deep learning-based multi-view structure-from-motion framework and state-of-the-art deep multi-view 3D reconstruction methods.
\begin{figure}[!ht]
\centering
\includegraphics[width=0.45\textwidth]{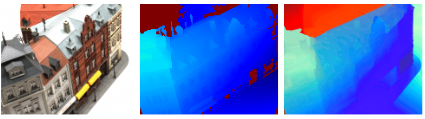}
\caption{The reference image (left), the ground truth depth map (center) and the output depth map (right).}
\label{fig:depth_images}
\end{figure}
\begin{figure}[!ht]
\centering
\includegraphics[width=0.45\textwidth]{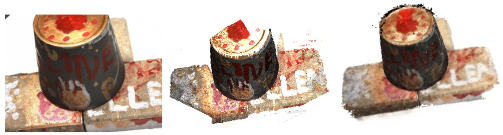}
\caption{The input image, ground truth, and output 3D reconstruction, respectively.}
\label{fig:pointcloud_images}
\end{figure}
\noindent
Figure \ref{fig:depth_images} shows an example of depth estimation and the qualitative comparison with the ground truth. The average error for the test dataset is 7.43, calculated from pixels with end-point-error greater than 3 pixels. Figure \ref{fig:pointcloud_images} shows the resulting 3D pointcloud and a qualitative comparison with  ground truth. Our method achieves an overall mean distance of 0.411. We compare with MVSNet, R-MVSNet and SurfaceNet on 17 scenes from the DTU dataset. As shown in Table \ref{tb:mvsnet}, our method achieves state-of-the-art performance in mean accuracy and mean completeness at faster computational times. 

\begin{table}[!ht]
\resizebox{0.49\textwidth}{!}{%
\begin{tabular}{lllll}
\hline
Method       & Mean Acc. & Mean Comp. & Overall (mm) & Runtime(s) \\ \hline
MVSNet       & 0.396     & 0.527      & 0.462        & 15.12      \\
R-MVSNet     & \textbf{0.385}     & 0.459      & 0.422        & 23.19      \\
SurfaceNet   & 0.450     & 1.04       & 0.745        & -          \\
Ours    & 0.391     &\textbf{ 0.429}        &   \textbf{0.411}                       & \textbf{2.19}       \\ \hline
\end{tabular}%
}
\caption{Mean Acc. is the mean accuracy of the distance metric(mm) and Mean Comp. is the mean completeness of the distance metric(mm). Runtime is the time of depth estimation for a pair of images.}
\label{tb:mvsnet}
\end{table}

\vspace{-20pt}
\section{Conclusion}
\vspace{-5pt}
In this paper we addressed the vision task of 3D reconstruction from images. Deep learning techniques have been shown to outperform procedural techniques. We incorporated a deep learning-based optical flow technique that uses hypercorrelation volumes to achieve accurate dense matching between images to a state-of-the-art two-view deep learning technique. We further extended this technique to the multi-view case. The evaluation on a complex benchmark dataset and further comparison with other state-of-the-art techniques shows that the proposed improvement is superior in terms of accuracy and performs faster. In the future, we plan on exploring alternative architectures for the reconstruction that would further improve the final accuracy.

\vspace{-10pt}
\section*{Acknowledgment}
\vspace{-5pt}
This research is based upon work supported by the Natural Sciences and Engineering Research Council of Canada Grants No.
N01670 (Discovery Grant) and DNDPJ515556-17 (Collaborative Research and Development with the Department of National
Defence Grant).


\bibliographystyle{abbrv}
\small 
\bibliography{main.bib}

\end{document}